\documentclass{article} 
\usepackage{iclr2020_conference,times}

\usepackage{amsmath,amsfonts,bm}

\def\eqref#1{equation~\ref{#1}}

\def\1{\bm{1}}

\DeclareMathAlphabet{\mathsfit}{\encodingdefault}{\sfdefault}{m}{sl}
\SetMathAlphabet{\mathsfit}{bold}{\encodingdefault}{\sfdefault}{bx}{n}

\usepackage{hyperref}
\usepackage{url}
\usepackage{booktabs}

\title{Spanish pre-trained BERT model \\ and evaluation data}

\iclrfinalcopy

\author{%
José Cañete\thanks{Work partially performed while at Adereso.}, Gabriel Chaperon, Rodrigo Fuentes \\
Department of Computer Science, Universidad de Chile \\
\texttt{\{jcanete,gchapero,rfuentes\}@dcc.uchile.cl} \\
\AND
Jou-Hui Ho, Hojin Kang \\
Department of Electrical Engineering, Universidad de Chile \\
\texttt{\{jouhui.ho,ho.kang.k\}@ug.uchile.cl} \\
\AND
Jorge Pérez \\
Department of Computer Science, Universidad de Chile \& \\
Millennium Institute for Foundational Research on Data (IMFD), Chile\\
\texttt{jperez@dcc.uchile.cl} \\
}

\begin{document}

\maketitle

\begin{abstract}
The Spanish language is one of the top 5 spoken languages in the world.
Nevertheless, finding resources to train or evaluate Spanish language models is not an easy task.
In this paper we help bridge this gap by presenting a BERT-based language model pre-trained exclusively on Spanish data.
As a second contribution, we also compiled several tasks specifically for the Spanish language in a single repository much in the spirit of the GLUE benchmark.  
By fine-tuning our pre-trained Spanish model, we obtain better results compared to other BERT-based models pre-trained on multilingual corpora for most of the tasks, even achieving a new state-of-the-art on some of them.
We have publicly released our model, the pre-training data, and the compilation of the Spanish benchmarks. 
\end{abstract}

\section{Introduction}
The field of natural language processing (NLP) has made incredible progress in the last two years.
Two of the most decisive features that have driven this improvement are the self-attention mechanism, particularly the Transformer architecture~\citep{vaswani2017attention}, and the introduction of unsupervised pre-training methods~\citep{peters2018deep,howard2018universal,devlin2018bert}, which take advantage of huge amounts of unlabeled text corpora.
Thus the leading strategy today for achieving good performance is to first train a Transformer-based model, say $M$, with a general language-modeling task over a huge unlabeled corpus and then, after this first training is over, ``fine-tune'' $M$ by continuing to train it for a specific task using labeled data. 
Built upon these ideas, the BERT architecture --which stands for ``Bidirectional Encoder Representations from Transformers''--~\citep{devlin2018bert}, and several improvements  thereof~~\citep{liu2019roberta,lan2019albert,yang2019xlnet,clark2019electra}, changed the landscape of NLP in 2019.


BERT was initially released in two versions, one pre-trained on an English corpus and another on a Chinese corpus~\citep{devlin2018bert}. 
As a way of providing a resource for other languages besides English and Chinese, the authors also released a ``multilingual'' version of BERT (we'll refer to it as mBERT from now on) pre-trained simultaneously on a corpus including more than 100 different languages.
The mBERT model has shown impressive performance when fine-tuned for language-specific tasks and has achieved state-of-the-art results in many cross-lingual benchmarks~\citep{wu2019beto,DBLP:journals/corr/abs-1906-01502}. 
The good performance of mBERT has drawn the attention of many different NLP communities, and efforts have been made to produce BERT versions trained on monolingual data.
This has led to the release of BERT models in Russian~\citep{kuratov2019adaptation}, French~\citep{martin2019camembert,le2019flaubert}, Dutch~\citep{de2019bertje,delobelle2020robbert}, Italian~\citep{PolignanoEtAlCLIC2019}, and Portugese~\citep{souza2019portuguese}.

In this paper we present the first BERT model pre-trained for the Spanish language.
Despite Spanish being widely spoken (much more than the previously mentioned languages), finding resources to train or evaluate Spanish language models is not an easy task.
For this reason, we also compiled several Spanish-specific tasks in a single repository, much in the spirit of the GLUE benchmark~\citep{wang2019glue}. 
By fine-tuning our pre-trained Spanish model, we obtain better results compared to other BERT-based models that were pre-trained on multilingual corpora for most of the tasks, and we even achieve a new state-of-the-art on some of them.
We have released our pre-trained model, the training corpus, and the compilation of benchmarks as free resources to be used by the Spanish NLP community.%
\footnote{\url{https://github.com/dccuchile/beto}}

In the rest of this paper, we first present our Spanish-BERT model. Then, we describe the tasks that we have compiled into a benchmark that we call GLUES (GLUE for Spanish), and finally, we present the results obtained by our model in some of the GLUES tasks.
Before going into the details of our model and results, we will briefly review the related work.


\section{Related Work}


Pre-trained language models using deep neural networks became very popular starting with  ULMFit~\citep{howard2018universal}.
ULMFit is based on a standard recurrent neural network architecture and a language-modeling task (predicting the next token from the previous sequence of tokens).
By using vast amounts of text, ULMFit is first trained for the language-modeling task, aiming to help the model acquire \emph{general knowledge} from a big corpus. 
The model is then fine-tuned in a supervised way to solve a specific task using labeled data.
The empirical results showed that the combination of pre-training and fine-tuning can considerably outperform training a model from scratch for the same supervised task.

A similar pre-training strategy was later used by~\cite{devlin2018bert}  to propose the BERT model. Compared with ULMFit, in BERT, the recurrent architecture is replaced with self-attention~\citep{vaswani2017attention}, which allows the prediction of a token to depend on every other token in the same sequence.
The task used for pre-training BERT, called \emph{masked language modeling}, is based on corrupting an input sequence by arbitrarily deleting some of the tokens and then training the model to reconstruct the original sequence~\citep{devlin2018bert}.
Several variations of BERT in terms of the training method and the task used for pre-training have been proposed~\citep{liu2019roberta,joshi2019spanbert, yang2019xlnet}. 
There have also been efforts to make models more efficient in terms of the number of parameters or training time~\citep{sanh2019distilbert, lan2019albert,clark2019electra}.




\citet{wu2019beto} and \citet{DBLP:journals/corr/abs-1906-01502} studied Multilingual BERT models, that is, models pre-trained simultaneously on corpora from different languages. 
These works showed how a single model can learn from several languages, setting strong baselines for tasks involving non-English languages.
XLM~\citep{lample2019cross} introduced a supervised objective which involved parallel multilingual data, and XLM-RoBERTa~\citep{conneau2019unsupervised} brought the multilingual models to the \emph{big leagues} in terms of model size. 

Several single-language BERT models came with results that usually got better performance than multilingual models as it is the case with 
CamemBERT~\citep{martin2019camembert} and FlauBERT~\citep{le2019flaubert} for French, BERTje~\citep{de2019bertje} and RobBERT~\citep{delobelle2020robbert} for Dutch, FinBERT~\citep{virtanen2019multilingual} for Finish, to name a few. 
Our work is similar to these models, but for the Spanish language. To the best of our knowledge, our paper presents the first publicly available Spanish BERT model and evaluation. 

\section{Spanish-BERT model, data and training}


We trained a model similar in size to a BERT-Base model~\citep{devlin2018bert}.
Our model has 12 self-attention layers, with 12 attention-heads each~\citep{vaswani2017attention}, using a hidden size of 768.
In total, our model has 110M parameters.
We trained two versions: one with cased data and one with uncased data, using a dataset that we will describe next.

For training our model, we collected text from different sources. 
We used all the data from Wikipedia and all the sources of the OPUS Project~\citep{tiedemann2012parallel} that had text in Spanish. 
These sources include United Nations and Government journals, TED Talks, Subtitles, News Stories, and more. 
The total size of the corpora gathered was comparable to the corpora used in the original BERT. 
Our training corpus contains about 3 billion words, and we have released it for later use.%
\footnote{\url{https://github.com/josecannete/spanish-corpora}}.
Our corpus can be considered an updated version of the one compiled by \cite{cardellinoSBWCE}.

For both versions of our model, cased and uncased, we constructed a vocabulary of 31K subwords using the byte pair encoding algorithm 
provided by the SentencePiece  library~\citep{kudo2018sentencepiece}. We added 1K placeholder tokens for later use, which gave us a vocabulary of 32K tokens.



For training our BERT models, we considered certain training details that have been successful in RoBERTa~\citep{liu2019roberta}.
In particular, we integrated the Dynamic Masking technique into our training, which involves using different masks for the same sentence in our corpus.
The Dynamic Masking we used was 10x, meaning that every sentence had 10 different masks. 
We also considered the Whole-Word Masking (WWM) technique from the updated version of BERT~\citep{devlin2018bert}. 
WWM ensures that when masking a specific token, if the token corresponds to a sub-word in a sentence, then all contiguous tokens conforming the same word are also masked.
Additionally, we trained on larger batches compared to the original BERT (but smaller than RoBERTa). 
We trained each model (cased and uncased) for 2M steps, with learning rate of $0.0001$, and the first $10000$ steps as warm-up.
The training was also divided into two phases, as described by \citet{you2019large}: we trained the first 900k steps with a batch size of 2048 and maximum sequence length of 128, and the rest of the training with batch size of 256 and maximum sequence length of 512. 
All the pre-training was done using Google's preemptible TPU v3-8.

\section{GLUES}
In this section, we present the GLUES benchmark, which is a compilation of common NLP tasks in the Spanish language, following the idea of the original English GLUE benchmark \citep{wang2019glue}. 
Through this benchmarks, we hope to help standardize future Spanish NLP efforts\footnote{See \url{https://github.com/dccuchile/glues} for a detailed description on how to obtain train-dev-test data for each task.}.
Next, we describe the tasks that we currently consider in GLUES.

\paragraph{Natural Language Inference: XNLI}
The Multi-Genre Natural Language Inference (MNLI) dataset \citep{williams2017broad} consists of pair of sentences. The first sentence is called the premise, and the second is the hypothesis. The task is to predict whether the premise entails the hypothesis (entailment), contradicts it (contradiction), or neither entails nor contradicts it (neutral). In other words, the task is a 3-class classification.
The Cross-Lingual Natural Language Inference (XNLI) corpus \citep{conneau2018xnli} is an evaluation dataset that extends the MNLI by adding development and test sets for 15 languages. 
In this setup, we train using the Spanish machine translation of the MNLI dataset, and use the development and test sets from the XNLI corpus. This task is evaluated using simple accuracy.

\paragraph{Paraphrasing: PAWS-X}

PAWS-X \citep{yang2019pawsx} is the multilingual version of the PAWS dataset \citep{zhang2019paws}. The task consists of determining whether two sentences are semantically equivalent or not.
The dataset provides standard (translated) train, development and test sets, and we used the Spanish portion. It is evaluated using simple accuracy.

\paragraph{Named Entity Recognition: CoNLL}
Named Entity Recognition consists of determining phrases in a sentence that contain the names of persons, organizations, locations, times, and quantities. 
For this task, we use the dataset by \citet{tjong-kim-sang-2002-introduction}, which focuses on persons, organizations, and locations, with a fourth category of miscellaneous entities, all tagged using the BIO format~\citep{BIO}.
This dataset provides standard train, development, and test sets, and the performance is measured with an F1 score.

\paragraph{Part-of-Speech Tagging: Universal Dependencies v1.4}
%
A part of speech (POS) is a category of words with similar grammatical properties, such as noun, verb, adjective, preposition, and conjunction.
For the POS tagging task, we use the Spanish subset of the Universal Dependencies (v1.4) Treebank \citep{udv1.4}. 
The version of the dataset was chosen following the works of \citet{wu2019beto} and \citet{kim-etal-2017-cross}.
The dataset provides standard train, development, and test sets. 
This task is evaluated based on the F1 score of predicted POS tags.

\paragraph{Document Classification: MLDoc}

The MLDoc dataset~\citep{SCHWENK18.658} is a balanced subset of the Reuters corpus. 
This task consists of classifying the documents into four categories, CCAT (Corporate/Industrial), ECAT (Economics), GCAT (Government/Social), and MCAT (Markets).
This dataset provides multiple sizes for the train split (1k, 2k, 5k and 10k), along with standard development and test sets for eight languages. We chose to train using the largest available train split in Spanish. This task is evaluated using simple accuracy.

\paragraph{Dependency Parsing: Universal Dependencies v2.2}
A dependency tree represents the grammatical structure of a sentence, defining relationships in the form of labeled-edges from ``dependent'' to ``head'' words. 
The label of each edge represents the type of relationship.
The task of dependency parsing consists of constructing a dependency tree for a given sentence.  
For this task, we use a subset of the Universal Dependencies v2.2 Treebank~\citep{udv2.2}, 
in particular, the concatenation of the AnCora and GSD Spanish portions of the dataset.
This decision and the version we chose follow the work of \citet{ahmad2018difficulties}.
The task is evaluated using the Unlabeled Attachment Score (UAS) and Labeled Attachment Score (LAS) \citep{DP}. UAS is the percentage of words that have been assigned the correct head, while LAS is the percentage of words that have been assigned both the correct head and the correct label for the relationship.



\paragraph{Question Answering: MLQA, TAR and XQuAD}
Given a context and a question, the task of question answering (QA) consists of finding the sequence of contiguous words within the context that answers the question.
This task is usually evaluated using two metrics averaged across all questions: exact match, that corresponds to the percentage of answers that match exactly, and F1 score, where answers are treated as bags of words. 
For our benchmark, we consider three translated versions of the SQuAD v1.1 dataset~\citep{rajpurkar2016squad}, namely, MLQA~\citep{lewis2019mlqa}, XQuAD~\citep{artetxe2019cross}, and TAR~\citep{carrino2019automatic}, which we will explain next.
MLQA provides translations for seven language with train data produced using machine translation from English, and development and test data translated by professionals~\citep{lewis2019mlqa}.
XQuAD provides test sets to evaluate cross-lingual models in 11 languages~\citep{artetxe2019cross}.
\citet{carrino2019automatic} proposed the TAR method (Translate Align Retrieve) that can be used to produce automatically machine translated versions of QA datasets. They provided a TAR-translation from English to Spanish of SQuAD v1.1~\citep{carrino2019automatic}. 
In our benchmark, we consider the train sets in MLQA and TAR, and the test sets in MLQA and XQuAD.






\section{Evaluation}
\subsection{Fine-tuning}
Since one of the goals of our work was to compare the performance of Spanish-BERT to mBERT~\citep{wu2019beto, yang2019pawsx}, our experimental setup closely follows the one by~\citet{wu2019beto}. 
Task specific output layers are incorporated, following the original BERT work~\citep{devlin2018bert}.

For each task, no preprocessing is performed except tokenization from words into subwords using the learned vocabulary and WordPiece~\citep{wordpiece}. 
We use the Adam optimizer~\citep{kingma2014adam} for fine-tuning with standard parameters ($\beta_1=0.9$, $\beta_2=0.999$), and L2 weight decay of $0.01$.  
We warm up the learning rate for the first 10\% of steps, with a linear decay afterward.

To allow for fine-tuning on a single GPU, we limit the length of each sentence to 128 tokens for all tasks. 
To accommodate tasks that require word-level classification, we use the sliding window approach described by \citet{wu2019beto}. After the first window, the last 64 tokens are kept for the next window, and only 64 new tokens are fed into the model.

For hyperparameter selection, we run experiments using different combinations of batch size, learning rate and number of epochs, following the values recommended by \citet{devlin2018bert}: batch size \{16, 32\}; learning rate \{5e-5, 3e-5, 2e-5\}; number of epochs \{2, 3, 4\}.
An extensive hyperparameter search is left for future work.

\begin{table}[t!]
\centering
\begin{tabular}{llllll}
\toprule
{\bf Model} & {\bf XNLI} & {\bf PAWS-X} & {\bf NER} & {\bf POS} & {\bf MLDoc} \\
\midrule
{Best mBERT} & 78.50$^a$ & 89.00$^b$ & {87.38}$^a$ & 97.10$^a$ & {95.70}$^a$  \\
\midrule
es-BERT uncased & 80.15 & \textbf{89.55} & 82.67 & {98.44} & \textbf{96.12$^*$}  \\
es-BERT cased & \textbf{82.01} & 89.05 & \textbf{88.43} & \textbf{98.97$^*$} & 95.60  \\  
\bottomrule
\end{tabular}
\caption{Comparison of Spanish BERT (es-BERT) with the best results obtained by  multilingual BERT models where the fine tune was done only on the Spanish train data in every dataset. 
Superscript denotes results obtained by: ($a$)~\citet{wu2019beto} and ($b$)~\citet{yang2019pawsx}.
The ``$^*$'' indicates a new state-of-the-art resut in the respective Spanish benchmark.
}
\label{tab:one}
\end{table}

\begin{table}[t!]
\centering
\begin{tabular}{llllllll}
\toprule
{\bf Model} & {\bf MLQA, MLQA} & {\bf TAR, XQuAD} & {\bf TAR, MLQA}\\
\midrule
{Best mBERT} & 53.90 / 37.40$^c$ & \textbf{77.60} / \textbf{61.80}$^d$ & 68.10 / \textbf{48.30}$^d$ \\
\midrule
es-BERT uncased & 67.85 / \textbf{46.03} & 77.52 / 55.46 & 68.04 / 45.00 \\
es-BERT cased &  \textbf{68.01} / 45.88 & 77.56 / 57.06 & \textbf{69.15} / {45.63}\\  
\bottomrule
\end{tabular}
\caption{Comparison of Spanish BERT (es-BERT) with the best results obtained by  multilingual BERT models in Question Answering. 
We only consider models where the fine tune was done on the Spanish train data in every dataset. Results are presented as F1 / ExactMatch. Every column header denotes the train set (left) and test set (right) used in everry case.
Different superscript denotes results obtained by different authors: 
($c$)~\citet{lewis2019mlqa} and 
($d$)~\citet{carrino2019automatic} }
\label{tab:two}
\end{table}

\subsection{Results}

Tables~\ref{tab:one} and~\ref{tab:two} show our results compared to the best mBERT result reported in the literature for the same setting. 
Spanish BERT outperforms most of the mBERT results, except for some QA settings (Table~\ref{tab:two}).
One of the largest difference can be seen in the XNLI task, which is the task that has the largest training dataset in Spanish. 
We note that for two of the most standard Spanish datasets (POS and MLDoc) we obtained a new state of the art.
We also note that there are some important differences in the performances in the QA datasets. 
The low quality of machine translation in MLQA might be one possible reason. 
We observed that nearly half of the 81K examples in MLQA have a mismatch between the answer and its starting position in the context.



It is important to note that our models are \emph{Spanish-only} and thus cannot take advantage neither of the original English train set in every translated benchmark, nor from train data in other languages. 
Taking advantage of data in different languages is a capability that multilingual models have by design.
In fact, there has been recent work on large multilingual models that can achieve better results on the Spanish datasets when fine-tuned with general, not necessarily Spanish, data. 
This is the case with the XLM-RoBERTa model~\citep{conneau2019unsupervised}, which, with 560M parameters and consuming training data from different languages, obtained results of 85.6\% for XNLI and 89\% for NER in the Spanish test set.
Both results are better than the one that we obtained by fine-tuning with Spanish data only (Table~\ref{tab:one}).
Similarly, \citet{yang2019pawsx} obtained 90.7\% in the Spanish test-set of PAWS-X when fine-tuning mBERT including the original English train set.



\section{Conclusion}
The advent of Transformer-based pre-trained language models has greatly improved the accessibility of high-performing models for the average user. In this paper, we successfully pre-train a Spanish-only model and open-source it, along with the training corpus and evaluation benchmarks, for the community to use. 

The ease of use of a pre-trained NLP model implies that its use cases are much broader, given that practitioners from disciplines other than computer science could fine-tune them for their domain-specific downstream tasks.
By releasing our Spanish-BERT model, we hope to encourage research and the application of deep learning techniques in Spanish-speaking countries.

Another direction of the work is to improve Spanish NLP models in terms of size, memory, and computation requirements. 
We are currently working on pre-training different ALBERT models \citep{lan2019albert} for Spanish, with number of parameters ranging from 5M to 223M. 
Our initial results with the smaller models are encouraging, and we plan to release these models as well. 


\subsubsection*{Acknowledgments}
We thank Adereso\footnote{\url{https://www.adere.so/}} for kindly providing support for training our uncased model, and Google for helping us with the Cloud TPU program for research. 
This work was partially funded by the Millennium Institute for Foundational Research on Data.

\bibliography{iclr2020_conference}
\bibliographystyle{iclr2020_conference}

\end{document}